\documentclass[letterpaper, 10 pt, conference]{ieeeconf}  

\IEEEoverridecommandlockouts                              

\title{\LARGE \bf
An End-to-End and Accurate PPG-based Respiratory Rate Estimation Approach Using Cycle Generative Adversarial Networks 
}

\author{Seyed Amir Hossein Aqajari$^{*,1}$, Rui Cao$^{1}$, Amir Hosein Afandizadeh Zargari$^{1}$, and Amir M. Rahmani$^{1,2,3}$
\thanks{ $^{1}$ Department of Electrical Engineering and Computer Science, University of California Irvine, CA 92697, USA
        {\tt\small (*correspondence e-mail: saqajari@uci.edu)}}%
\thanks{ $^{2}$ Department of Computer Science, University of California, Irvine, CA 92697, USA
        }
        
\thanks{ $^{3}$ Institute for Future Health, University of California, Irvine, CA 92697, USA
       }

}

\usepackage{graphicx}
\usepackage{amsmath}
\usepackage{subcaption}
\usepackage{siunitx}

\begin{document}

\maketitle
\thispagestyle{empty}
\pagestyle{empty}

\begin{abstract}


Respiratory rate (RR) is a clinical sign representing ventilation. An abnormal change in RR is often the first sign of health deterioration as the body attempts to maintain oxygen delivery to its tissues. There has been a growing interest in remotely monitoring of RR in everyday settings which has made photoplethysmography (PPG) monitoring wearable devices an attractive choice.
PPG signals are useful sources for RR extraction due to the presence of respiration-induced modulations in them. The existing PPG-based RR estimation methods mainly rely on hand-crafted rules and manual parameters tuning. An end-to-end deep learning approach was recently proposed, however, despite its automatic nature, the  performance  of  this  method  is  not  ideal using the real world data. In this paper, we present an end-to-end and accurate pipeline for RR estimation using Cycle Generative Adversarial Networks (CycleGAN) to reconstruct respiratory signals from raw PPG signals. Our results demonstrate a higher RR estimation accuracy of up to 2$\times$ (mean absolute error of 1.9$\pm$0.3 using five fold cross validation) compared to the state-of-th-art using a identical publicly available dataset. Our results suggest that CycleGAN  can be a valuable method for RR estimation from raw PPG signals.


\end{abstract}

\section{INTRODUCTION}
Respiratory rate (RR), often referred to as breathing rate, is the number of breaths a person takes per minute. A normal resting RR for adults ranges from 12 to 20 \cite{clevelandclinic}. Abnormal changes in respiratory rate are an accurate indicator of physiological conditions such as anxiety, hypoxia, hypercapnia, metabolic and respiratory acidosis \cite{rolfe2019importance}. A diverse body of research studies has indicated the significance of respiration rate for forecasting events such as cardiac arrest, patient deterioration, and care escalation \cite{cooper2014respiratory,dougherty2015royal,cretikos2008respiratory}.

The importance of respiratory rate as one of the first indicators of health deterioration has attracted significant attention in RR's daily monitoring \cite{nicolo2020importance}. However, RR's reliable measurement devices are bulky and cumbersome, and are mainly used for inpatients. With the rapid development of wearable technologies, a change in an individual's physiological systems' functional state can be tracked and monitored in an everyday setting, for instance, by using photoplethysmography (PPG) \cite{allen2007photoplethysmography}. PPG signals can easily be collected continuously and remotely using a wide range of inexpensive, convenient, and portable wearable devices (.e.g., smart watches, rings, etc.). The blood perfusion dynamics are known to carry information on breathing, as respiration-induced modulations in PPG signals \cite{pirhonen2018acquiring}. Hence, PPG signals are considered as a suitable source for respiratory rate extraction to forecast unexpected care admissions in a daily life setting .

The RR estimation from PPG signals has received remarkable attention in the literature \cite{charlton2017breathing}. Traditional RR estimation methods require several steps, including digital filtering, time/frequency domain analysis, extraction of signal components from composite signals, deriving respiratory surrogate waveforms and features using the fiducial points, signal quality estimations, and sensor fusion \cite{charlton2016assessment}. These techniques rely heavily on manual parameter tuning, optimization, and hand-crafted rules designed for specific target patient population. In the past few years, machine learning techniques and neural networks have been widely used in health monitoring domains \cite{zargari2020newertrack, aqajari2021pyeda, mehrabadi2020detection, aqajari2021pain, zargari2021accurate}. Bian et al. \cite{bian2020respiratory} recently proposed an end-to-end deep learning approach in order to automatically and accurately estimate RR from raw PPG signals. Despite the automatic nature of their proposed model, the performance of this method is not ideal (mean absolute error (MAE) of 3.8 $\pm$ 0.5 bpm, which is about \%25 inaccuracy considering 16 bpm as an average RR per minute). 


This work proposes an automatic end-to-end generative deep learning approach using cycle generative adversarial networks (CGAN) \cite{CycleGAN2017} to reconstruct respiratory signals from raw PPG signals and estimate RR with a high accuracy. CGAN is a novel and powerful approach in the field of unsupervised learning, which targets learning the structure of two given data domains to translate an individual input from one domain to a desired output from the second domain. We also propose a novel loss function to be integrated in our CGAN model that takes into account the key attribute (i.e., RR) of the generated respiratory signals. 
Our results demonstrate that the proposed GAN-based approach estimates RR from raw PPG signals with 2$\times$ higher accuracy compared with the state-of-the-art approach \cite{bian2020respiratory} using real-world data. Furthermore, our method outperforms the classical RR estimations methods, despite utilizing the complete automatic end-to-end design. 

In summary, this work makes the following key contributions:
\begin{itemize}
    \item Proposing an end-to-end automatic approach based on CGAN which outperforms the performance of the classical RR estimation methods (using an identical setting and dataset).
    \item Proposing a novel loss function for our CGAN model that takes into account the RR of the generated respiratory signals.
    \item Demonstrating the performance of our approach using the real-world data and comparing it against the state-of-the-art (using an identical setting and dataset).

\end{itemize}


The  rest  of  this  paper  is  organized  as  follows. Section \ref{methods} introduces the employed dataset and our proposed pipeline architecture. In Section \ref{results} we summarize the result obtained by our proposed method. Section \ref{discussion} compares our result with the state-of-the-art in RR estimation from raw PPG signal. Finally, Section \ref{conclusion} concludes the paper.

\section{MATERIAL AND METHODS}
\label{methods} 

\subsection{Dataset}

We employed BIDMC PPG and Respiration Dataset \cite{pimentel2016toward} to evaluate our RR estimator method. This is a publicly available dataset which contains signals and numerics extracted from the much larger MIMIC II matched waveform Database, along with manual breath annotations made from two annotators, using the impedance respiratory signal.

In this dataset, PPG and impedence respiratory signals are collected from 53 adult patients for about 8-minute duration at sampling rate of 125 Hz. This dataset is widely used to evaluate the performance of different algorithms for estimating respiratory rate from PPG signals \cite{charlton2017breathing, birrenkott2017robust, jarchi2018accelerometry, bian2020respiratory}.

\subsection{RR Estimation Pipeline}

\begin{figure}[h]
  \centering
  \includegraphics[width=0.8\linewidth]{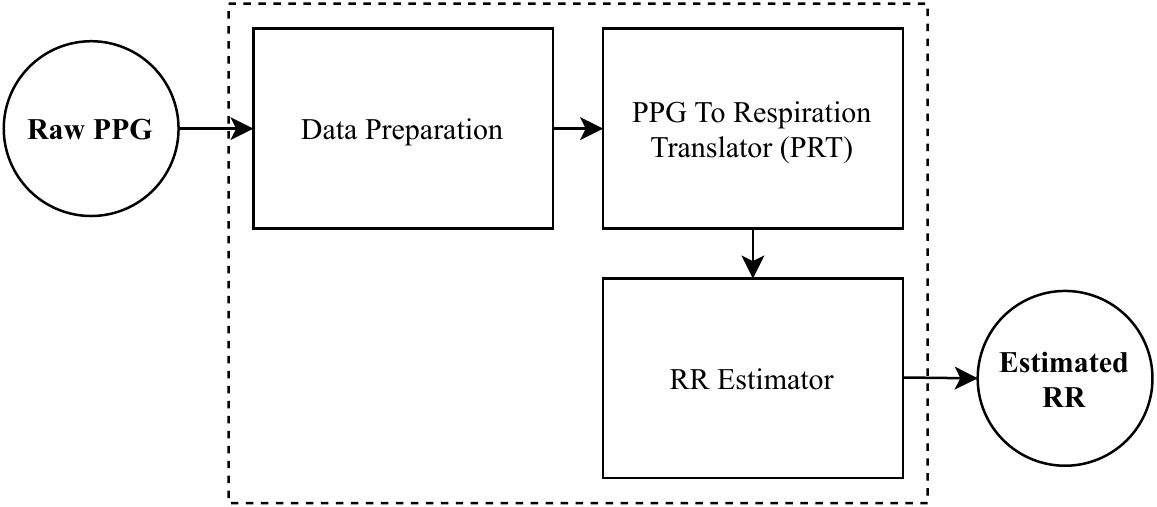}
  \caption{RR Estimation Pipeline}
  \label{fig:pipeline}
\end{figure}

Figure \ref{fig:pipeline} shows our proposed pipeline for estimating respiratory rate from PPG signals. There are three different main stages in this pipeline: (1) Data Preparation, (2) PPG to Respiration Translator (PRT), and (3) RR Estimator. In the following subsections, we discuss each part in detail.
\subsubsection{Data Preparation} 
The primary purpose of this stage is to prepare the data and pre-process it for the PPG-to-Respiration Translator (PRT) module. PPG signals are sampled at much higher frequency than required in the BIDMC dataset. Therefore, downsampling is done to save memory, and reduce processing time and computational complexity of our model while preserving the signals integrity. In this stage, first, raw PPG data are normalized to 0-1. Then, the signals are down-sampled to 30 Hz. Finally, 30-second windows of data are extracted from the signals to be used in the PRT module for translation. 

\subsubsection{PPG to Respiration Translator (PRT)}
In this module, the Cycle Generative Adversarial Networks (The Cycle GAN) are employed to reconstruct respiratory signals from raw PPG signals. The Generative Adversarial Networks belong to the field of unsupervised learning targeting to learn the structure of a given data in order to generate new unseen data. The GANs are composed of two models: a generator network and a discriminator network. The generator network starts at a point from a latent space as an input and aims to generate new data similar to the expected domain. The discriminator network on the other hand attempts to recognize if an input data is real (belongs to the original dataset) or fake (generated by the generator network).

The Cycle GAN is an extension of the Generative Adversarial Networks which was first proposed by Jun-Yan Zhu et al. \cite{CycleGAN2017}. The idea behind the Cycle GAN is to take an input from the first domain and generate an output of the second domain. In our case, the goal of Cycle GAN is to learn the mapping between PPG signals (domain \textit{X}) and respiratory signals (domain \textit{Y}). Each domain contains set of training samples $\{{x_i}\}_{i=1}^N \in X$ and $\{{y_i}\}_{i=1}^N \in Y$ used directly from BIDMC dataset. The model includes two generators with mapping functions as $G : X \rightarrow Y$ and $F : Y \rightarrow X$ and two discriminators $D_X$ and $D_Y$. In the discriminator networks, $D_X$ aims to distinguish between real PPG signals ($x_i$) and synthetic PPG signals ($F(y)$) while $D_Y$ aims to discriminate between real respiratory signals ($y_i$) and synthetic respiratory signals ($G(x)$).

We indicate the distributions of our data as $x \sim p_{data}(x)$ and $y \sim p_{data}(y)$. Our objective loss function contains three terms: (1) adversarial losses \cite{goodfellow2014generative}, (2) cycle consistency losses, and (3) RR Loss.

Adversarial losses are employed for matching the distribution of generated synthetic signals to the data distribution of original signals. We apply adversarial loss function on both of our mapping functions. The objective function applied to the mapping function \textit{G} is expressed as below:

\begin{align}
\begin{split}
L_{GAN}(G, D_Y, X, Y) = E_{y\sim p_{data}(y)}[log D_{Y}(y)]\\  + E_{x\sim p_{data}(x)}[log(1- D_{Y}(G(x)))]
\end{split}
\end{align}

where $G$ tries to generate respiratory signals $(G(x))$ that look similar to original respiratory signals collected from BIDMC dataset (domain Y), while $D_Y$ aims to discriminate between synthetic respiratory signals $(G(x))$ and real samples $(y)$. In a same way, adversarial loss for the mapping function \textit{F} is expressed as $L_{GAN}(F, D_X, Y, X)$.

A mapping function trained only by adversarial loss as an objective function can map the same set of signals from the first domain to any random permutation of signals in the second domain. Therefore, cycle consistency losses are added to guarantee the mapping from an individual input ($x_i$) to a desired output $(y_i)$ by considering learned mapping functions to be cycle consistent. This means that for each PPG signal $x$ from domain X we must have $x \rightarrow G(x) \rightarrow F(G(x)) \approx x$ while for each respiratory signal $y$ we have $y \rightarrow F(y) \rightarrow G(F(y)) \approx y$. This behaviour is indicated as:

\begin{align}
\begin{split}
L_{cyc}(G, F) = E_{x\sim p_{data}(x)}[||F(G(x))-x||_1]\\  + E_{y\sim p_{data}(y)}[||G(F(y))-y||_1]
\end{split}
\end{align}

In order to force the synthesized respiratory signals to only keep their main features, we define RR loss function which attempts to take into account the RR of the generated respiratory signals. The BioSPPy \cite{biosppy} public python library is used to calculate the respiration rate of the synthetic and original respiratory signals. This additional loss function can be expressed as:

\begin{align}
L_{RR}(G) = E_{y\sim p_{data}(y)}[||G(F(y))_{RR}-y_{RR}||_1]
\end{align}

Therefore, the final objective is the weighted sum of the above loss functions:

\begin{align}
\begin{split}
L(G, F, D_X, D_Y) = L_{GAN}(G, D_Y, X, Y) \\ + L_{GAN}(F, D_X, Y, X) \\ + \lambda_1 L_{cyc}(G, F) \\ + \lambda_2 L_{RR}(G)
\end{split}
\end{align}

where $\lambda_1$ and $\lambda_2$ are the weights of cycle consistency loss and RR loss respectively (both are empirically selected as 10 in our work). 

$G$ and $F$ attempt to minimize this objective against adversaries $D_X$ and $D_Y$ that try to maximize it. Hence, we aim to solve:

\begin{align}
G^*,F^* = \arg \min_{G, F} \max_{D_X, D_Y}L(G, F, D_X, D_Y)
\end{align}

We use the CGAN architecture proposed by \cite{CycleGAN2017}. The architecture of generative networks is adopted from Johnson et al. \cite{johnson2016perceptual}. This network contains two stride-2 convolutions, several residual blocks \cite{he2016deep}, and two fractionally-strided convolutions with stride 0.5. For the discriminator networks we use 70$\times$70 PathGANs \cite{isola2017image, li2016precomputed, ledig2017photo} which aims to classify whether the signals are fake or real.

\subsubsection{RR Estimator}
This module uses the BreathMetrics \cite{noto2018automated} for pre-processing, filtering, and calculating the respiratory rate of the signals. The BreathMetrics is a Matlab toolbox for analyzing respiratory recordings. Breathmetrics was designed to make analyzing respiratory recordings easier by automatically de-noising the data and extracting the many features embedded within respiratory recordings including respiratory rate. The methods used in this tool (de-noising and features extraction) have been validated, peer-reviewed, and published in Chemical Senses, a scientific journal. 


\begin{figure}[h]
  \centering
  \includegraphics[width=0.5\linewidth,height=0.7\linewidth]{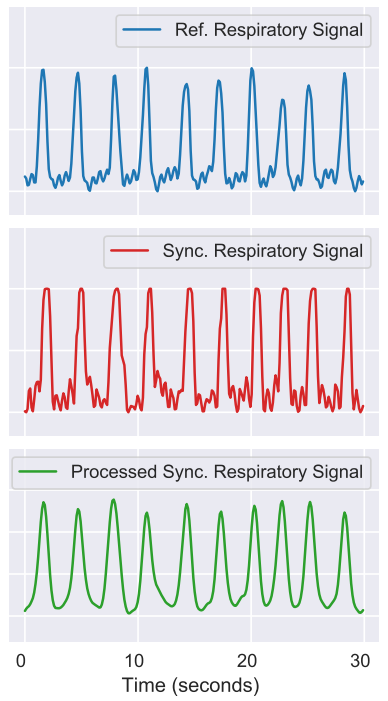}
  \caption{Example of the synthetic respiratory signal generated by our PRT module, from top to bottom: reference respiratory signal, synthetic respiratory signal, and processed synthetic respiratory signal using the BreathMetrics module.}
  \label{fig:sample}
\end{figure}

\subsection{Performance Metric}

We calculate the mean absolute error (MAE) as the performance metric in our study in order to evaluate our RR estimation method. MAE is calculated by averaging the absolute differences between the values estimated by a model and the values observed.

MAE is defined as:


\begin{align}
\begin{split}
MAE = \frac{1}{N}\sum_{i=1}^{N} |RR_{e}^i-RR_{r}^i| \\ 
\end{split}
\end{align}


Where $N$ is total number of respiratory segments. $RR_{e}^i$ and $RR_{r}^i$ are estimated and reference respiratory rate for each 1 minute data respectively.

To evaluate the performance of our algorithm, we use 5-fold cross validation. We split the BIDMC dataset into five folds while making sure each subject's data is appeared in one fold only. We perform the training 5 times and each time 4 folds are for training and 1 fold is for testing. The average MAE of 5 runs is presented as the final performance results for our proposed model. We trained our models on Nvidia Quadro RTX 5000 GPU with 125 GB of RAM. For each training experiment we performed 100 epochs in which early stopping technique were employed to reduce over-fitting.

\section{RESULTS}
\label{results}

Figure \ref{fig:sample} shows an example of a reconstructed respiration signal using our method alongside with its reference respiration signal for the same timing window of the same test subject. The blue signal represents the reference respiration signal from the BIDMC dataset. The red signal is the synthetic respiratory signal which is an output of our PRT module. The green signal shows the final respiratory signal after de-noising and pre-processing of the signal using the BreathMetrics tool. By comparing the reference respiratory signal and the processed synthetic respiratory signal it can be seen that our pipeline is capable of reconstructing the respiratory signal itself from the PPG signal with high precision.

Table \ref{tbl:features} shows the summary of the MAE performance (average$\pm$std) of our method compared with the state of the arts. The same publicly available dataset (BIDMC) and the same performance metric (MAE using five fold cross validation) are used to fairly evaluate and compare the performance of our automatic approach.


\begin{table}[!h]
\centering
\caption{The summary of MAE Performances in compare with state of the arts}
\label{tbl:features}
\resizebox{.8\columnwidth}{!}{%
\begin{tabular}{|c|c|c|}
\hline
 \textbf{RR Estimation Models} & \textbf{MAE}\\
\hline
Bian et al. \cite{bian2020respiratory} DL method & 3.8$\pm$0.5\\
Bian et al. \cite{bian2020respiratory} SmartQualityFusion method & 2.6$\pm$0.4\\
Our proposed CycleGAN-based method & 1.9$\pm$0.3\\
\hline
\end{tabular}
}
\end{table}

As can be seen from the table, our CGAN-based method significantly (up to 2$\times$) outperforms the state-of-the-art RR estimation methods from a raw PPG signals. In the next section, we discuss the results in more details.

\section{DISCUSSION}
\label{discussion} 

Existing PPG-based RR estimation methods heavily rely on features derived from various hand-crafted algorithms and tuning parameters for specific settings. In this work, we presented a novel fully end-to-end automatic approach for estimating RR from raw PPG signals.

A classical RR estimation method has been implemented in \cite{bian2020respiratory} which is referred as SmartQualityFusion, as a combination of Smart fusion \cite{karlen2013multiparameter} and Quality fusion \cite{chan2013ambulatory}. The SmartQualityFusion algorithm achieved the MAE of 2.6$\pm$0.4 breaths/min (Table \ref{tbl:features}). According to Table \ref{tbl:features}, our proposed RR estimation pipeline demonstrate a significant improvement in accurately estimating RR compared with their algorithm (the MAE of 1.9 breaths/min), despite the complete end-to-end automatic environment of our method.


Bian et al. \cite{bian2020respiratory}, also proposed an end-to-end learning approach based on deep learning to estimate RR from PPG. Their results demonstrate a clinically reasonable performance; however, the performance of their proposed automatic approach heavily depends to the availability of real world data. They used the mean absolute error (MAE) as a performance metric in their work.  According to their results, their deep learning approach trained with real data could achieve the MAE of 3.8$\pm$0.5 breath/minutes which is substantially higher than the MAE error achieved by their classical algorithm (SmartQualityFusion).
According to Table \ref{tbl:features}, our proposed pipeline architecture 
provides about 2 times better performance of estimating respiratory rate compared with their proposed deep learning approach. 


One existing limitation of our work, which is also present in the other state-of-the-art methods, is the lack of noisy PPG data in the real datasets used for training our models. Therefore, the performance of our proposed method might not be as expected on a noisy PPG data collected during daily life and physical activities. This limitation is also present in the other state-of-the-arts methods since the employed datasets are mostly consist of stationary data. However, as also mentioned in \cite{bian2020respiratory}, this problem can easily be diminished by retraining the model using the new existing noisy PPG and respiratory signals. 

In \cite{bian2020respiratory}, the authors increased the performance of their proposed method, by augmenting the real datasets using the generated synthetic PPG data. This enhanced the MAE of their model from 3.8 brpm to 2.5 brpm. Thus, the data augmentation approach significantly enhanced the performance of their model. However, our proposed model still achieves a better performance (1.9 brpm), despite being trained only on a real data. As a future work, we intend to further increase the performance of our proposed method by augmenting our dataset using synthetic data. 

\section{Conclusion}
\label{conclusion}
In summary, in this work we presented a novel pipeline architecture in order to estimate respiratory rate using PPG signals. We are the first one to use cycle adversarial networks in our model to reconstruct the respiratory signals from PPG signals. According to our results, our proposed pipeline architecture is able to estimate RR with the MAE of 1.9 which has the performance of 2.0x better than the state-of-the-art automatic RR estimation method.






\bibliographystyle{IEEEtran}
\bibliography{IEEEabrv,ref}

\end{document}